\documentclass[10pt,twocolumn,letterpaper]{article}

\usepackage[pagenumbers]{cvpr} 

\usepackage{adjustbox}

\definecolor{cvprblue}{rgb}{0.21,0.49,0.74}
\usepackage[pagebackref,breaklinks,colorlinks,allcolors=cvprblue]{hyperref}
\usepackage[capitalize]{cleveref}

\usepackage{listings}
\lstdefinestyle{pythonstyle}{
    language=Python,
    basicstyle=\ttfamily\small,
    keywordstyle=\bfseries\color{blue},
    stringstyle=\color{red},
    commentstyle=\color{gray},
    morecomment=[l][\color{magenta}]{\#},
    numbers=left,
    numberstyle=\tiny\color{gray},
    stepnumber=1,
    numbersep=10pt,
    backgroundcolor=\color{white},
    showspaces=false,
    showstringspaces=false,
    showtabs=false,
    frame=single,
    tabsize=4,
    captionpos=b,
    breaklines=true,
    breakatwhitespace=false,
    morekeywords={as,assert,async,await,break,continue,def,del,elif,except,finally,from,global,import,lambda,nonlocal,pass,raise,try,with,yield},
    escapeinside={(*@}{@*)},
}

\usepackage{etoc}
\etocdepthtag.toc{mtchapter}
\etocsettagdepth{mtchapter}{subsection}
\etocsettagdepth{mtappendix}{none}

\RequirePackage{xspace}
\makeatletter
\DeclareRobustCommand\onedot{\futurelet\@let@token\@onedot}
\def\@onedot{\ifx\@let@token.\else.\null\fi\xspace}
\def\aka{\emph{a.k.a}\onedot}

\def\ie{\emph{i.e}\onedot}

\def\etal{\emph{et al}\onedot}

\crefname{theorem}{Theorem}{Theorem}
\crefname{lemma}{Lemma}{Lemma}
\crefname{remark}{Remark}{Remark}
\crefname{figure}{Fig.}{Fig.}
\crefname{section}{Sec.}{Sec.}
\crefname{equation}{Eq.}{Eq.}
\crefname{table}{Tab.}{Tab.}
\crefname{algorithm}{Alg.}{Alg.}

\newcommand{\myPara}[1]{\vspace{.05in}\noindent\textbf{#1}}

\def\paperID{8201} 
\def\confName{CVPR}
\def\confYear{2025}

\title{ScaMo: Exploring the Scaling Law in Autoregressive Motion Generation Model
}

\author{Shunlin Lu\textsuperscript{123}, Jingbo Wang\textsuperscript{3}\dag, Zeyu Lu\textsuperscript{5}, Ling-Hao Chen\textsuperscript{4}, Wenxun Dai\textsuperscript{4}, \\Junting Dong\textsuperscript{3}, Zhiyang Dou\textsuperscript{6}, Bo Dai\textsuperscript{36}\dag, Ruimao Zhang\textsuperscript{1}\\
\\
\textsuperscript{1} Sun Yat-sen University  \textsuperscript{2} The Chinese University of Hongkong, Shenzhen \\ \textsuperscript{3} Shanghai AI Laboratory  \textsuperscript{4} Tsinghua University  \\ \textsuperscript{5} Shanghai Jiao Tong University  \textsuperscript{6} 
The University of Hong Kong
}

\begin{document}
\twocolumn[{%
\renewcommand\twocolumn[1][]{#1}%
\maketitle
\vspace{-3em}
\begin{center}
    \centering
    \captionsetup{type=figure}
    \includegraphics[width=\textwidth]{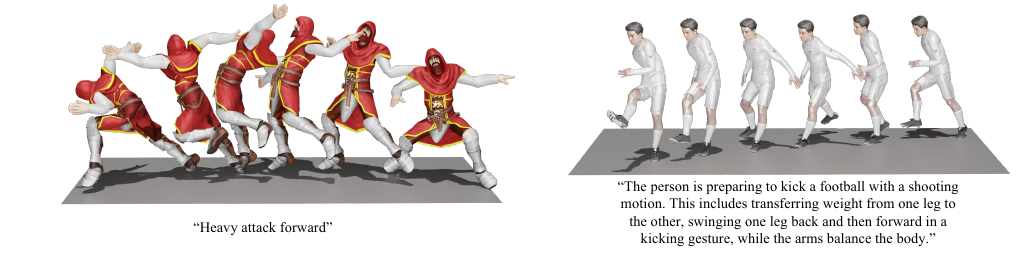}
    \vspace{-2.5em}
    \captionof{figure}{The generation results of ScaMo-3B with a text input. Our model could deal with abstract sentences and long sentences.}
    \label{fig:visualization}
\end{center}%
}]
\let\thefootnote\relax\footnotetext{$^\dagger$ Corresponding author
}
\begin{abstract}

The scaling law has been validated in various domains, such as natural language processing (NLP) and massive computer vision tasks; however, its application to motion generation remains largely unexplored. In this paper, we introduce a scalable motion generation framework that includes the motion tokenizer Motion FSQ-VAE and a text-prefix autoregressive transformer. Through comprehensive experiments, we observe the scaling behavior of this system. For the first time, we confirm the existence of scaling laws within the context of motion generation. Specifically, our results demonstrate that the normalized test loss of our prefix autoregressive models adheres to a logarithmic law in relation to compute budgets. Furthermore, we also confirm the power law between Non-Vocabulary Parameters, Vocabulary Parameters, and Data Tokens with respect to compute budgets respectively. Leveraging the scaling law, we predict the optimal transformer size, vocabulary size, and data requirements for a compute budget of $1e18$. The test loss of the system, when trained with the optimal model size, vocabulary size, and required data, aligns precisely with the predicted test loss, thereby validating the scaling law. Project page: \href{https://shunlinlu.github.io/ScaMo/}{https://shunlinlu.github.io/ScaMo/}
\end{abstract}
\section{Introduction}
\label{sec:intro}

Scaling law gives the precise prediction of test loss, optimal model size, and data requirements given a specific compute budget in FLOPs. This capability enables researchers to conduct relatively small-scale experiments and accurately forecast the performance on larger scales, thereby conserving research time and compute resources. In recent years, the scaling law has been extensively studied, particularly in the field of natural language processing (NLP), where many large language models (LLMs)~\citep{kaplan2020scaling, gpt4, bai2023qwen, bi2024deepseek} have empirically validated their scaling properties.

\begin{figure}[tb] \centering
\includegraphics[width=1\linewidth]{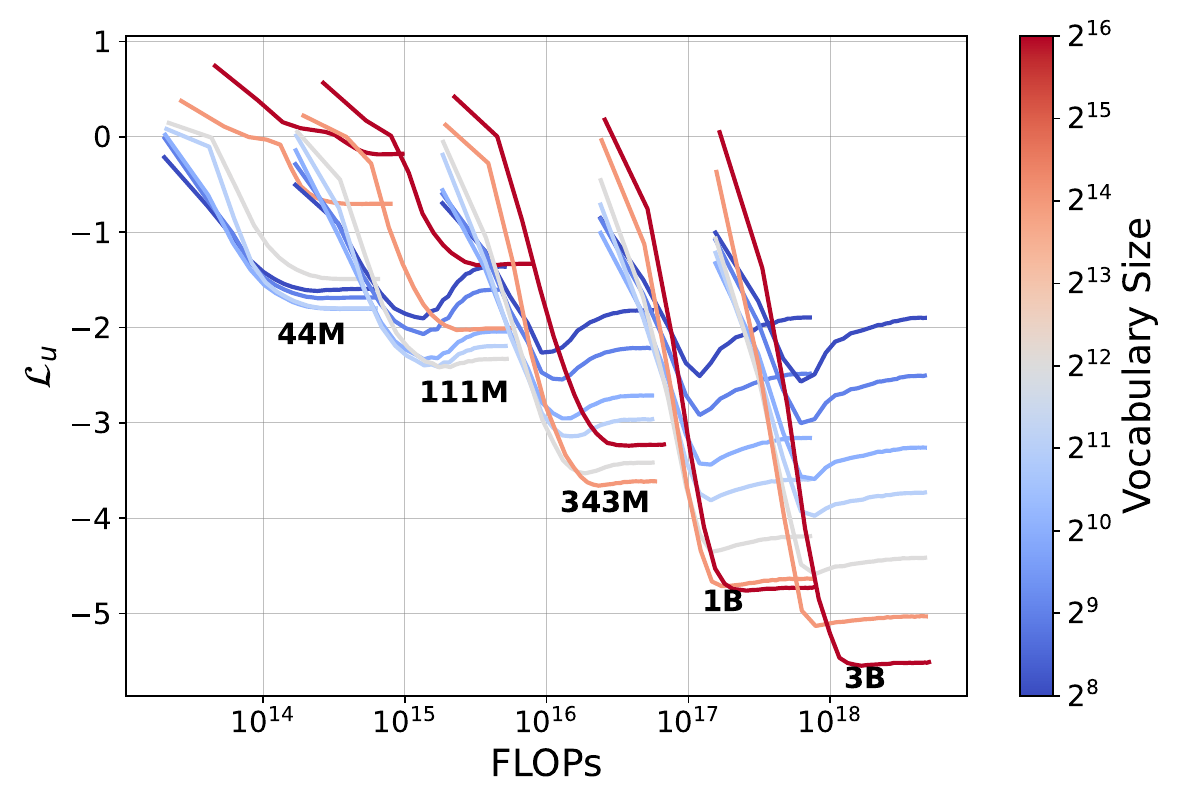}
    \vspace{-1.8em}
    \caption{We plot the relationship between normalized test loss and FLOPs for observing the scaling behavior. Overall, the larger model and larger vocabulary size can get better performances.} 
    \label{fig:teaser_all_loss}
    \vspace{-1em}
\end{figure}

Building on these prior successes, recent research has extended scaling laws to the community of computer vision, particularly for tasks such as text-to-image generation~\citep{liang2024scaling, sun2024autoregressive}. However, the scaling properties within the realm of human motion generation remain relatively underexplored. This is largely due to the challenges associated with the costly process of motion data collection and the substantial computational resources required. To guide the allocation of data collection efforts and compute budgets for further training, our objective is to examine whether similar scaling behaviors can be observed in motion generation tasks. Specifically, we take a transformer-based decoder-only auto-regressive motion generation framework as an example. To reach a targeted test loss, we need to set the computation budget accordingly, which can further be used to determine the optimal data requirements, appropriate model parameters, and vocabulary size.

Exploring the scaling law of the auto-regressive model is natural and applicable in text-driven motion generation.
However, building a motion generation framework at a larger scale has still not been explored well in the community. In this work, we mainly try to answer a research question, \textit{What hinders the verification of scaling law in text-driven motion generation?} To this end, we attempt to answer this question from the following aspects.

\begin{figure*}
	\centering
    \captionsetup[subfigure]{aboveskip=-2pt, belowskip=0em}
	\begin{subfigure}{0.495\linewidth}
		\centering
		\includegraphics[width=1.0\linewidth]{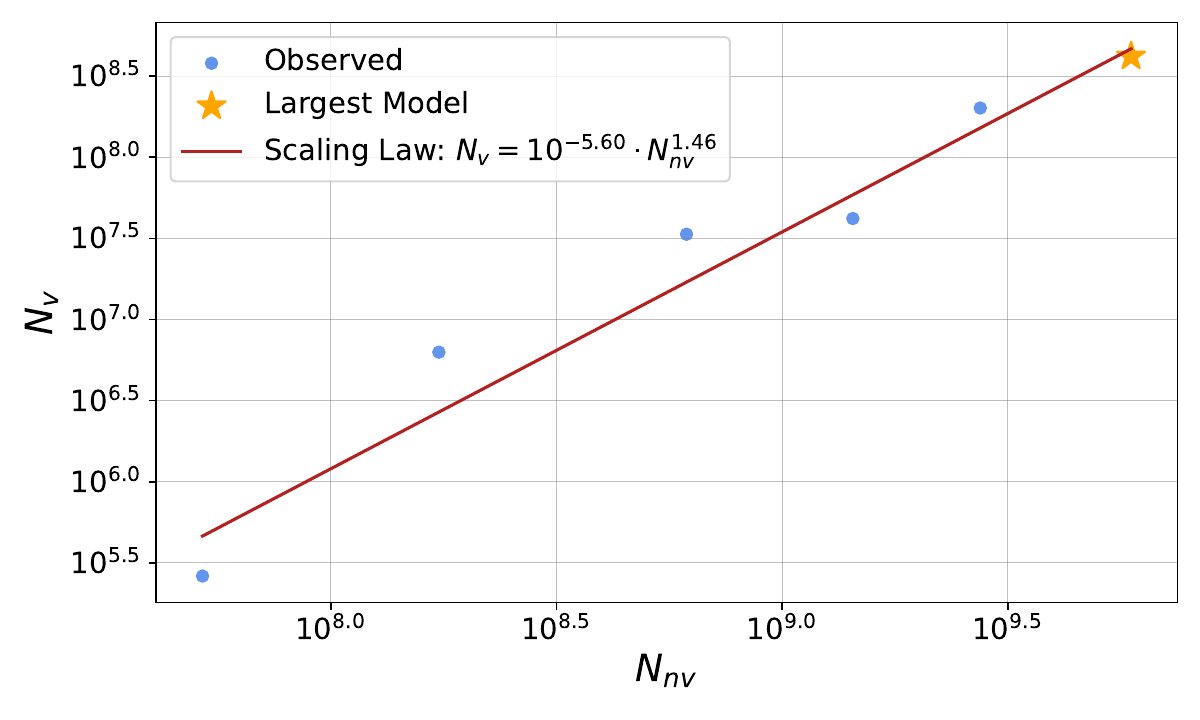}
		\caption{Non-Vocabulary Parameter Scaling Law.}

	\end{subfigure}
	\begin{subfigure}{0.495\linewidth}
		\centering
		\includegraphics[width=1.0\linewidth]{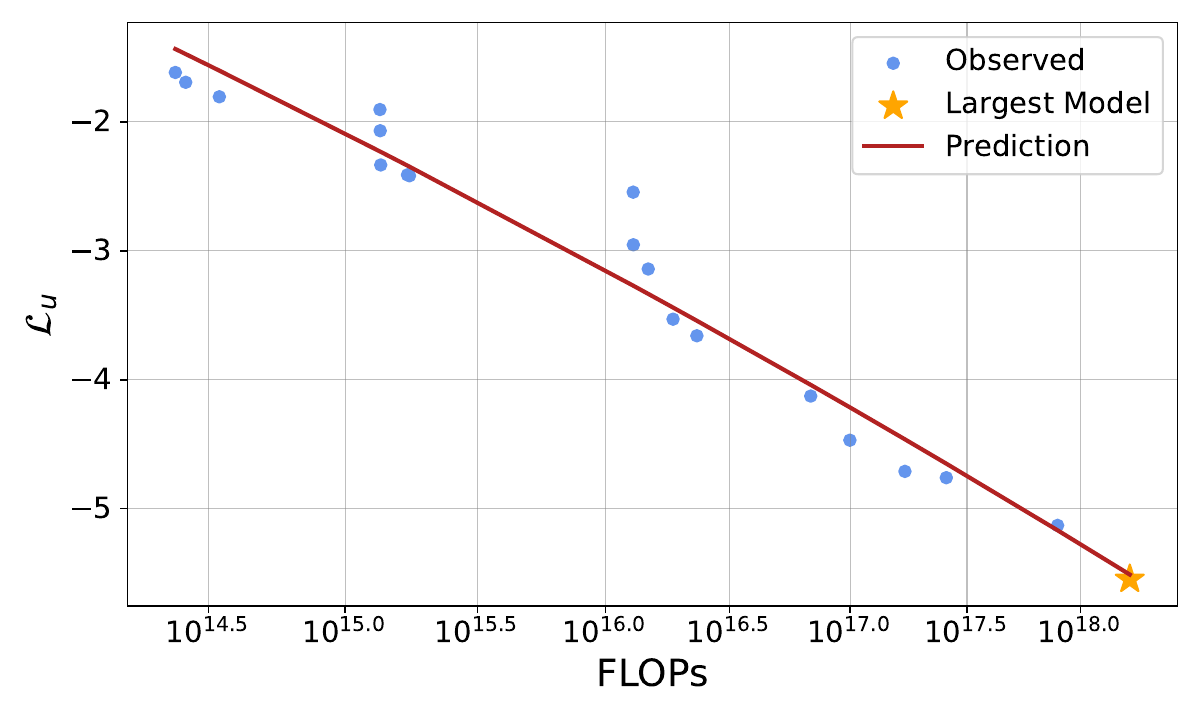}
		\caption{Performance Scaling Law.}

	\end{subfigure}
        \vspace{-1em}
	\caption{Scaling laws of ScaMo. (a) Power law between $N_{nv}$ and $N_v$. We could predict the $Nv$ precisely based on a given $N_{nv}$. (b) Logarithmic law between FLOPs $C$ and normalized test loss $\mathcal{L}_u$. We could predict the $\mathcal{L}_u$ precisely given a FLOPs $C$.}
        \vspace{-1.5em}
	\label{fig:scaling law}
\end{figure*}

(i) \textbf{Limited data scale and quality.} The scale of motion data is much less than languages, images, or videos, due to its expensive collection process. Specifically, the largest used dataset Motion-X~\citep{motionx} only contains 98,000 sequences, which is not enough for observing the scaling properties. A concurrent work~\citep{wang2024quo} collected a dataset of over 1M motion sequences. However, in this dataset, over half of the sequences consist of one-frame pose repeated 64 times, introducing a significant number of static motions.

(ii) \textbf{Difficulties in scaling the vocabulary size of tokenizer.} In the auto-regressive motion generation framework, directly scaling the size of the motion codebook is almost in vain. In contrast to tokenizers used in text, image, and video modalities, the vocabulary in motion tokenization is insufficient for na\"ive scaling. Directly applying existing vector quantization (VQ) methods for tokenization fails to scale effectively. When the codebook size increases, VQ suffers from codebook collapse, resulting in low utilization of codebooks. Therefore, exploring an effective tokenizer for auto-regressive human motion generation is urgent.

(iii) \textbf{Insufficient scalability of the model architecture.} Previous work tries to introduce a large language model with an extended vocabulary~\citep{motiongpt, wang2024quo}, which compromises the generation performance with diverse downstream tasks, such as motion understanding. Furthermore, directly introducing the LLMs into motion generation makes it hard to explore the scalability of model size, due to the limited size choices of pretrained foundational LLMs. Other works train autoregressive models from scratch~\citep{t2mgpt}. They use the sentence embedding from CLIP~\citep{clip}. However, recent work~\cite{motionclr} points out that the sentence-level language guidance for motion generation is not fine-grained enough for cross-modality alignment. Therefore, how to marriage foundational text encoders with a large motion generation model is still not well studied in the community.

To resolve the limitations above, we present a scalable system, denoted as ScaMo, to better investigate the scaling properties in text-driven motion generation. Unlike previous works directly using the existing dataset, we first collect a dataset of over 260 hours of motion data from sources such as Motion-X~\citep{motionx}, CombatMotion~\citep{CombatMotion}, 100-Style~\citep{100style}, and an internal dataset. This dataset, denoted as MotionUnion, does not suffer from the static motion issue that was observed in previous studies~\citep{wang2024quo}. For the motion tokenization process, we adopt a more generalizable approach, finite scale quantization (FSQ)~\citep{FSQ}, only relying on a simple reconstruction loss. This method can maintain performance similar to Vanilla VQ~\citep{t2mgpt} at a relatively small codebook size. In contrast, when scaling with a larger codebook size, FSQ mitigates the codebook collapse issue and achieves superior performance. Additionally, to avoid compromising the language compression capability of the model, we use a frozen T5-XL~\cite{t5} as the encoder. Importantly, the text is encoded as word-level embeddings prefixed in the auto-regressive motion generation model. In this transformer-based motion generator, we apply bidirectional attention to text tokens and causal attention to motion tokens. The text tokens are visible to all motion tokens. These improvements enable us to perform a comprehensive study on the scalability of motion generation models. We plot the relationship between test normalized loss and FLOPs in~\cref{fig:teaser_all_loss}. Especially, we observe the scaling behaviors for the first time in motion generation.

Our experiments reveal several important insights: \textbf{Logarithmic law relationship between normalized test loss and FLOPs}. We observe a logarithmic relationship between the normalized test loss and computational resources (FLOPs). From this, we can predict the achievable test loss for a given FLOPs. In the absence of computational constraints, larger FLOPs should be preferred to maximize performance. This implies that larger vocabulary sizes and larger transformers should be selected. The larger vocabularies make the model more expressive, and a larger vocabulary requires a larger transformer.  \textbf{Power law between vocabulary size, model size, and data token with respect to FLOPs.} With the given FLOPs, we can predict the optimal vocabulary size, model size and data tokens to get the best performance. Furthermore, we find the power relationship between non-vocabulary parameters and vocabulary parameters. Our prediction also suggests that vocabulary parameters should be scaled faster than non-vocabuolary parameters, i.e., $N_v \propto N_{nv}^{\gamma}$, where $\gamma\approx1.46 > 1$, shown in~\cref{fig:scaling law}.

To verify the effectiveness of the estimated scaling law, we set a fixed computation budget to $1\times 10^{18}$. With the obtained scaling law, we predict the optimal model size, optimal vocabulary size, and required data are set as $3B$, $2^{16}$, and $1\times10^{7.5}$ respectively. When training a model in this setting, we found that the test normalized loss precisely aligns with the predicted loss, as shown in Fig~\ref{fig:scaling law}. When training on our dataset, the model enjoys smooth generalization capabilities for diverse text inputs and generates higher-quality motions, whose results are shown in~\cref{fig:visualization}. We believe our model shows significant potential for scaling, with the capacity to improve motion generation quality on larger text-motion datasets. Additionally, it exhibits the ability for out-of-distribution text inputs.

Our contribution can be summarized as (1) We first demonstrate the existence of scaling laws in the motion generation task from a practical perspective. (2) We have revealed the core factors that limit the scaling laws in motion generation of previous work are the lack of data and unscalable model architectures. Our proposed scalable motion generation system opens the door to scalable research in motion generation. (3) The trained models can deal with complex sentences and generate more vivid motions.

\begin{figure}[tb] \centering
    \includegraphics[width=0.9\linewidth]{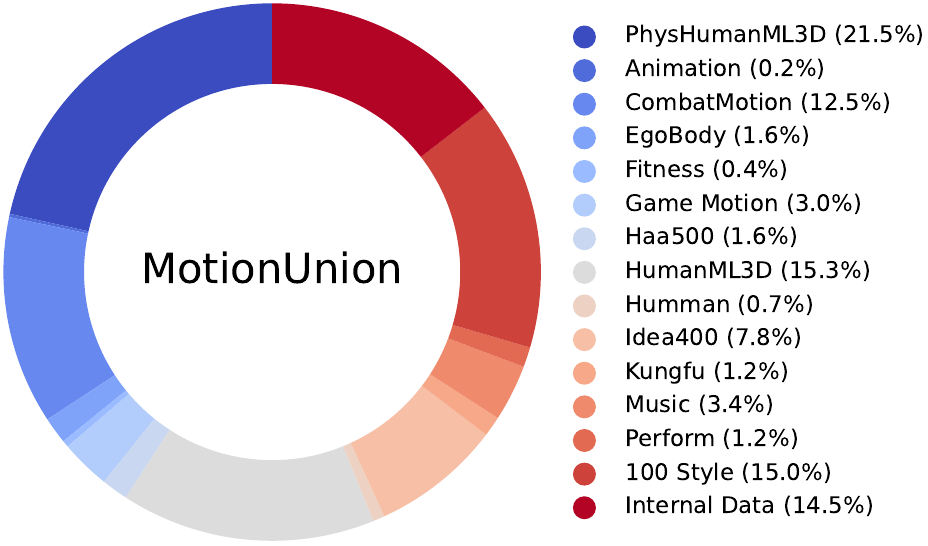}
        \vspace{-0.7em}
    \caption{The frames statistics of MotionUnion dataset. Motion capture data accounts for the majority.}
        \vspace{-0.8em} 
        \label{fig:data}
\end{figure}

\section{Related Work}
\label{sec:related}
\myPara{Text-aligned human motion generation}~\citep{plappert2018learning, text2action, dvgans,  jl2p, t2g, motionclip, temos, avatarclip, tm2t, motiondiffuse, teach, mdm, humanise, mld, mofusion, physdiff, t2mgpt, diffprior, remodiffuse, gmd, motionclr, motiongpt, motiongptv2, unihsi, omnicontrol, humantomato, tlcontrol, momask, promotion, amd, b2ahdm, emdm, stmc, flowmdm, move, stablemofusion,zhang2024generative} develop fast in recent years, owing to both the progresses in generative models~\cite{ddim,ddpm1,ddpm2,attention} and the scales of datasets~\cite{motionx,interx,humanml3d}. For methods, the introduction of GPT-like~\citep{t2mgpt, humantomato, momask, motiongpt} motion generation method and the diffusion-based method~\citep{mdm, motiondiffuse, remodiffuse,emdm,mld,motionlcm} boost the development of human motion generation in a large extend. For datasets, KIT~\cite{kit} and HumanML3D~\cite{humanml3d} are two representative datasets supporting human motion with open vocabulary inputs. Recent progress~\cite{motionx,interx,omg} in human motion generation tries to enhance the generation quality via scaling the size of datasets. However, how to synthesize human motion with more data, larger models and motion tokens is still under-explored.

\myPara{The scaling law in generative models} has attracted significant attention in recent years mainly due to the significant progress in auto-regressive language generation~\cite{gpt3,gpt4,llama}. Besides, the progress in the visual generation community~\cite{tian2024visual,sora,dalle2,ldm} also verifies the effectiveness of scaling data, model size, and computation. Exploring the relationship between these resources and the model performance is extremely essential to predict the experimental results when scaling on resource-cost scenarios. Early research~\cite{banko2001scaling} has verified the power scaling law with the data scale. In recent years, some research in the language community~\cite{hestness2019beyond,hestness2017deep,kaplan2020scaling} additionally investigates the scaling law with the model and data. Hoffmann \etal~\cite{hoffmann2022training} show how to scale with the data and the model size jointly. Recent study~\cite{tian2024visual} also verifies the power scaling law of auto-regressive models in image generation with model size and computational cost. However, how to scale with data, model size, and vocabulary size in the auto-regressive (sequential) motion generation process is still under exploration.

\begin{figure*}[tb] 
    \centering    
    \includegraphics[width=0.9\linewidth]{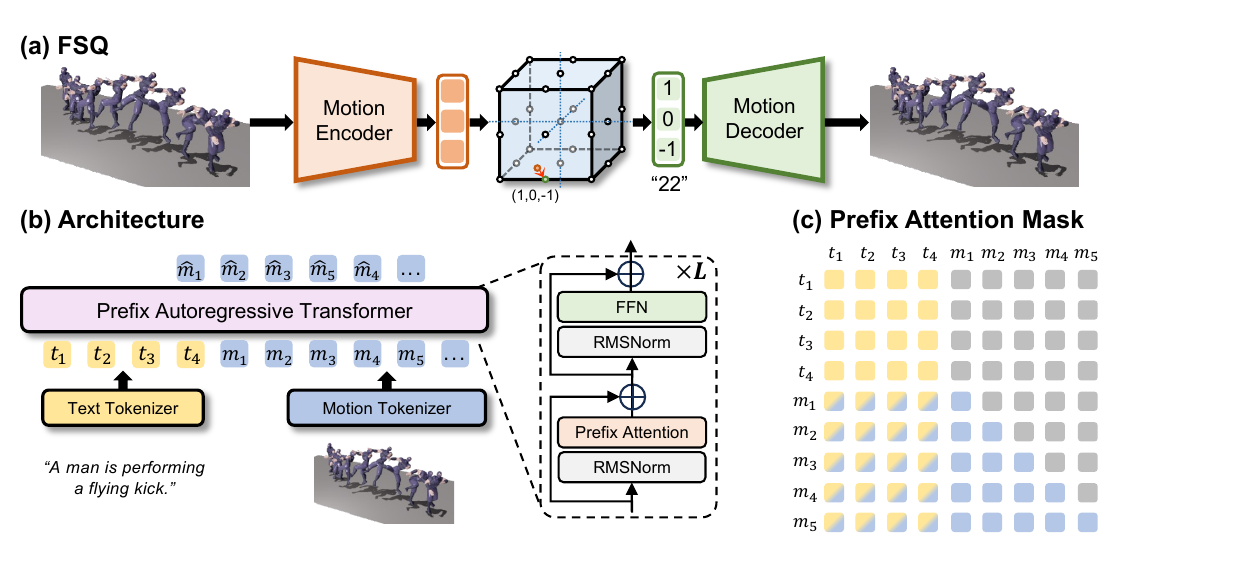}
    \vspace{-1em}
    \caption{Overview of ScaMo architecture. \textbf{(a) FSQ}: Motion FSQ-VAE. We use one code quantization and $d=L=3$ as an example. The feature of other frames is quantized in the same way. \textbf{(b) (c) Text-prefix Autoregressive Transformer}: The text tokens are applied with bidirectional attention and the motion tokens are applied with causal attention. Motion tokens can attend all text tokens.} 
    \vspace{-0.5em}
    \label{fig:pipeline}
\end{figure*}

\section{Dataset Construction}
Previous research~\cite{t2mgpt} indicates that the existing dataset suffers from server overfitting, which limits its utility for evaluating scaling properties. To better explore these properties, we introduce a new, large-scale text-motion dataset, MotionUnion, comprising approximately 150k sequences and 30M frames. Detailed statistics are provided in the Appendix. Each sequence in the dataset is paired with a corresponding sentence. The dataset includes data from several sources, such as Motion-X~\citep{motionx}, Combatmotion~\citep{CombatMotion}, 100 Style~\citep{CombatMotion}, and Physics-corrected HumanML3D, along with additional internal data. The internal data are sourced from manually created animations or motion capture, and are annotated with text generated by GPT-4. Following the methodology outlined in Smoodi~\citep{zhong2024smoodi}, we retarget the raw motion data to the SMPL skeleton and employ the same processing pipeline as HumanML3D~\citep{humanml3d} to obtain the motion representations. Examples of the data can be found in the Appendix. The dataset composition and the proportion of each component are illustrated in \cref{fig:data}.

\vspace{-1em}
\section{Scalable Motion Generation System }

Motion is inherently suitable for autoregressive modeling following tokenization. Similarly to previous work, our scalable framework comprises two key components: a motion tokenizer and a prefix autoregressive generation model, as depicted in \cref{fig:pipeline}. In the realms of image and video processing, the pursuit of large vocabulary sizes has become commonplace. To accurately capture motion details, a large vocabulary size is also essential the field of motion generation.  However, the advantages of employing a large vocabulary size have not been widely recognized yet. Previous studies have not demonstrated notable improvements or robustness when utilizing large vocabularies. Upon further investigation, we hypothesize that this limitation arises from the codebook collapse phenomenon induced by the quantization method. To address this issue, we propose a scalable finite scalar quantizer, which is discussed in~\cref{sec:4.1}. Additionally, to enhance text encoding, we introduce the prefix autoregressive model in~\cref{sec:4.2}.

\subsection{Finite Scalar Quantization}
\label{sec:4.1}
\myPara{Vanilla motion VQ-VAE.}  
Motion VQ-VAE is designed to learn discrete representations of human motion sequences through an encoding-decoding framework. Specifically, the model employs a vector quantization (VQ) mechanism to reconstruct motions, where an auto-encoder architecture is utilized. The model learns a codebook $\mathcal{C} = \{\mathbf{e}_k\}_{k=1}^K$, with $K$ representing the size of the codebook and $\mathbf{e}_k$ denoting the $k$-th embedding in the codebook.

Given a latent vector $\mathbf{z}$ and a quantizer $\mathcal{Q}(\cdot; \mathcal{C})$, the quantization process selects the codebook entry $\hat{\mathbf{z}}$ that minimizes the reconstruction error relative to $\mathbf{z}$, formally defined as,
\vspace{-0.7em}
\begin{equation}
    \hat{\mathbf{z}} = \mathcal{Q}(\mathbf{z} ; \mathcal{C})=\underset{\mathbf{e}_k}{\arg \min }\|\mathbf{z}-\mathbf{e}_k\|_{2}^{2}. 
    \vspace{-0.7em}
\end{equation}

In the context of vanilla Motion VQ-VAE, the latent vector $\mathbf{z}$ is derived from the motion encoder $\mathtt{Enc}(\cdot)$ applied to a motion sequence $\mathbf{m} \in \mathbb{R}^{T \times D}$, where $T$ is the number of motion frames and $D$ is the feature dimension of each frame. The motion is further reconstructed through the decoder $\mathtt{Dec}(\cdot)$, with the overall model optimized to minimize the following loss function,
\vspace{-0.5em}
\begin{equation}
    \label{eq:vq}
    \mathcal{L} = \|\mathbf{m} - \mathtt{Dec}(\mathcal{Q}(\mathbf{z}; \mathcal{C}))\|_{2}^{2} + \alpha \|\mathbf{z} - \mathtt{sg}(\hat{\mathbf{z}})\|_{2}^{2},
    \vspace{-0.5em}
\end{equation}
where $\alpha$ is a hyperparameter, $\mathtt{sg}(\cdot)$ denotes the stop-gradient operation, and $\mathtt{Dec}(\cdot)$ is the motion decoder.

Unlike traditional VQ-VAE frameworks, the codebook $\mathcal{C}$ in Motion VQ-VAE is updated using two techniques: exponential moving average (EMA) and codebook reset, as proposed in T2M-GPT~\citep{t2mgpt}. While the use of discrete vector quantization in vanilla VQ-VAE effectively compresses human motion data, the quantization error remains significant. To address this issue, one might consider increasing the size of the codebook. However, a direct increase in codebook size can lead to codebook collapse, resulting in degraded performance and instability during training, especially as the model scales to handle more complex motion data, which is shown in~\cref{sec:5.2}.

\myPara{Motion FSQ-VAE.} 

To resolve the issue of codebook collapse, we analyze the key problem that lies in the ``$\arg\min$'' operation of vanilla VQ-VAE. The matching process is done by comparing the distance between $\mathbf{z}$ and $\hat{\mathbf{z}}$ via ``$\arg\min$'' operation, leading the optimizer to prefer the specific parts of the codebook and ignore updating others. To resolve this issue, we use a finite scalar quantizer (FSQ) instead. Technically, instead of using $\arg\min$, FSQ tries to round $\mathbf{z}$ as,
\vspace{-0.5em}
\begin{equation}
    \hat{\mathbf{z}} = \mathcal{Q}(\mathbf{z}) = \mathtt{round}(f(\mathbf{z})),
    \vspace{-0.5em}
\end{equation}
where $f(\cdot)$ is the bounding function, setting as the $\mathtt{sigmoid}(\cdot)$ function in our practice. Each channel in $z$ will be quantized into one of the unique $L$ integers, therefore we have $\hat{z} \in \{1,\ldots,L\}^d$. The codebook size is calculated as $|\mathcal{C}| = \prod_{i=1}^{d} L_i$. Here, $L$ is a super parameter, and we follow Mentzer \etal~\citep{FSQ} to set $L_i \geq 5$. We leave the $L_i$ we use in the appendix. 
Similarly, the $\mathtt{round}(\cdot)$ operation can not propagate gradients, thus the stop-gradient technique is used. The optimization objective is only the reconstruction loss is required without other tricks, \ie,
\vspace{-0.3em}
\begin{equation}
    \mathcal{L} = \|\mathbf{m} - \mathtt{Dec}(f(z) + \mathtt{sg}(\mathtt{round}(f(z))-f(z)))\|_{2}^{2},
    \vspace{-0.3em}
    \label{eq:fsq}
\end{equation}
where the encoders and decoders are adapted from vanilla VQ-VAE~\citep{t2mgpt}. Note that FSQ is a replacement for VQ and can also be extended to group FSQ or residual FSQ. We leave it as our further work.

\subsection{Text Prefix Autoregressive Model}
\label{sec:4.2}
We revisit how the previous motion generator integrates with the foundational language models. We notice that directly extending the vocabulary of language to motion tokens is not effective enough for text-driven motion generation. Differently, we introduce the world-level language prefix for the generation process. In contrast to some classical auto-regressive motion generation models~\cite{t2mgpt}, the input sentence is encoded as tokens for each word. As illustrated in~\cref{fig:pipeline}, the attention calculation within the word part is bidirectional while the attention calculation for the motion part is causal attention. In this way, we could leverage the text embedding from the frozen text encoder, and the prefixed auto-regressive model is only optimized by the motion tokens part using the cross-entropy loss,
\vspace{-0.3em}
\begin{equation}
    \label{eq:ce}
    \mathcal{L} = - \sum_{t=1}^{n} \log p(\hat{m}_t | m_{<t}, S, V),
\vspace{-0.4em}
\end{equation}
where $S$ denotes the text and $V$ denotes the vocabulary.

\subsection{Scaling Law Formulation in Our Model}
\label{sec: scaling law in our model}

Classical scaling law only supports the fixed vocabulary, thus we follow the previous work \cite{scalingvocab} to reformulate. With the given compute budget \( C \) in FLOPs, the goal is to get the optimal non-vocabulary model parameters \( N_{nv} \), the optimal vocabulary model parameters \( N_v \), the number of training tokens \( D \). \( N_v \) is calculated as \(N_v = Vd\), where \(V\) is the vocabulary size and $d$ is the dimension of the model. Accordingly, these terms can be formulated as, 
\vspace{-0.3em}
\begin{equation}
\begin{aligned}
(N_v^{\text{opt}}, N_{nv}^{\text{opt}}, D^{\text{opt}}) = 
&\ \arg \min_{N_v, N_{nv}, D} \mathcal{L}(N_v, N_{nv}, D) \\ &\ \text{s.t.} \quad \text{FLOPs}(N_v, N_{nv}, D) = C,
\end{aligned}
\vspace{-0.5em}
\label{equ1}
\end{equation}

According to Kaplan \etal~\citep{kaplan2020scaling}, the FLOPs (\(C\)) of our Transformer-based model can be estimated as,
\vspace{-0.5em}
\begin{equation}
\begin{aligned}
C \approx 6ND = 6(N_v+N_{nv})D \approx 6(N_{nv} + Vd)D.
\end{aligned}
\vspace{-0.5em}
\label{equ2}
\end{equation}
To capture the scaling law in text-driven motion generation, we follow Hoffmann \etal~\citep{hoffmann2022training} to train models with different model sizes from 44M-3B and vocabulary sizes from $2^8$ to $2^{16}$. Since the model with a large vocab size naturally has a higher cross-entropy loss. We follow Tao \etal~\citep{scalingvocab} to use the normalized loss for fair evaluation, 
\vspace{-0.5em}
\begin{equation}
\begin{aligned}
\mathcal{L}_u = - \frac{1}{T} \sum_{t=1}^{T} \mathtt{\log} \frac{p(m_t | m_{< t}, S, V)}{p(m_t | S, V)},
\label{eq:normalized_loss}
\end{aligned}
\vspace{-0.7em}
\end{equation}
Then we plot the IsoFLOPs figure, shown in~\cref{fig:teaser_all_loss}.
Similarly to previous work, we hypothesize the power law equations can capture the relationship between quantities. We formulate the relationship as, 
\vspace{-0.7em}
\begin{equation}
\begin{aligned}
N_{v}^{\text{opt}} \propto C^a , \quad N_{nv}^{\text{opt}} \propto C^b, \quad \text{and} \quad D^{\text{opt}} \propto C^c.
\end{aligned}
\vspace{-0.5em}
\end{equation}

We choose the pre-defined FLOPs that can reach the lowest normalized loss and fit the above power law equations to obtain the above coefficients.

\begin{figure}
	\centering
    \captionsetup[subfigure]{aboveskip=0pt, belowskip=0pt}
	\begin{subfigure}{1.0\linewidth}
		\centering
		\includegraphics[width=1.0\linewidth]{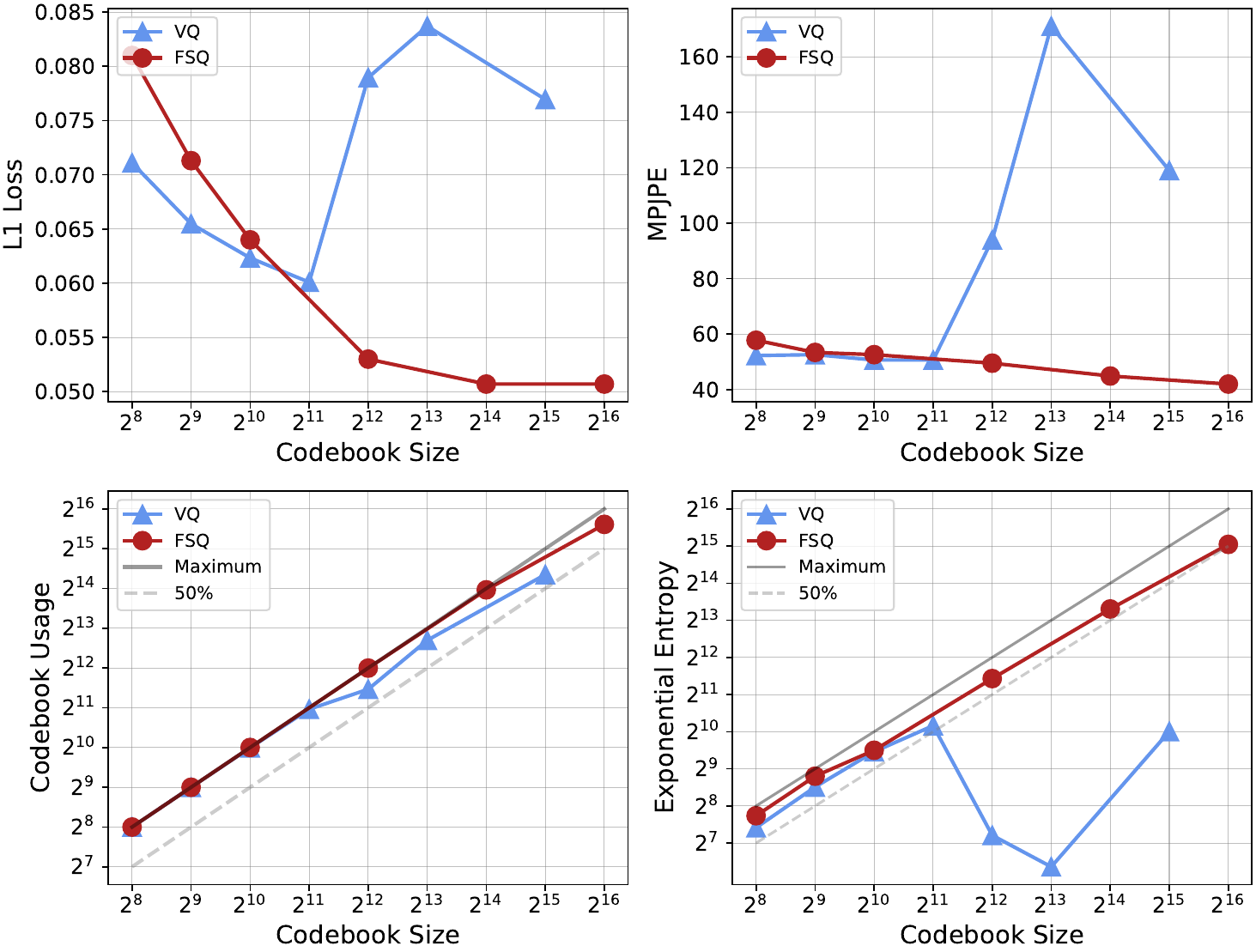}
		\caption{HumanML3D.}
	\end{subfigure}
	\begin{subfigure}{1.0\linewidth}
		\centering
		\includegraphics[width=1.0\linewidth]{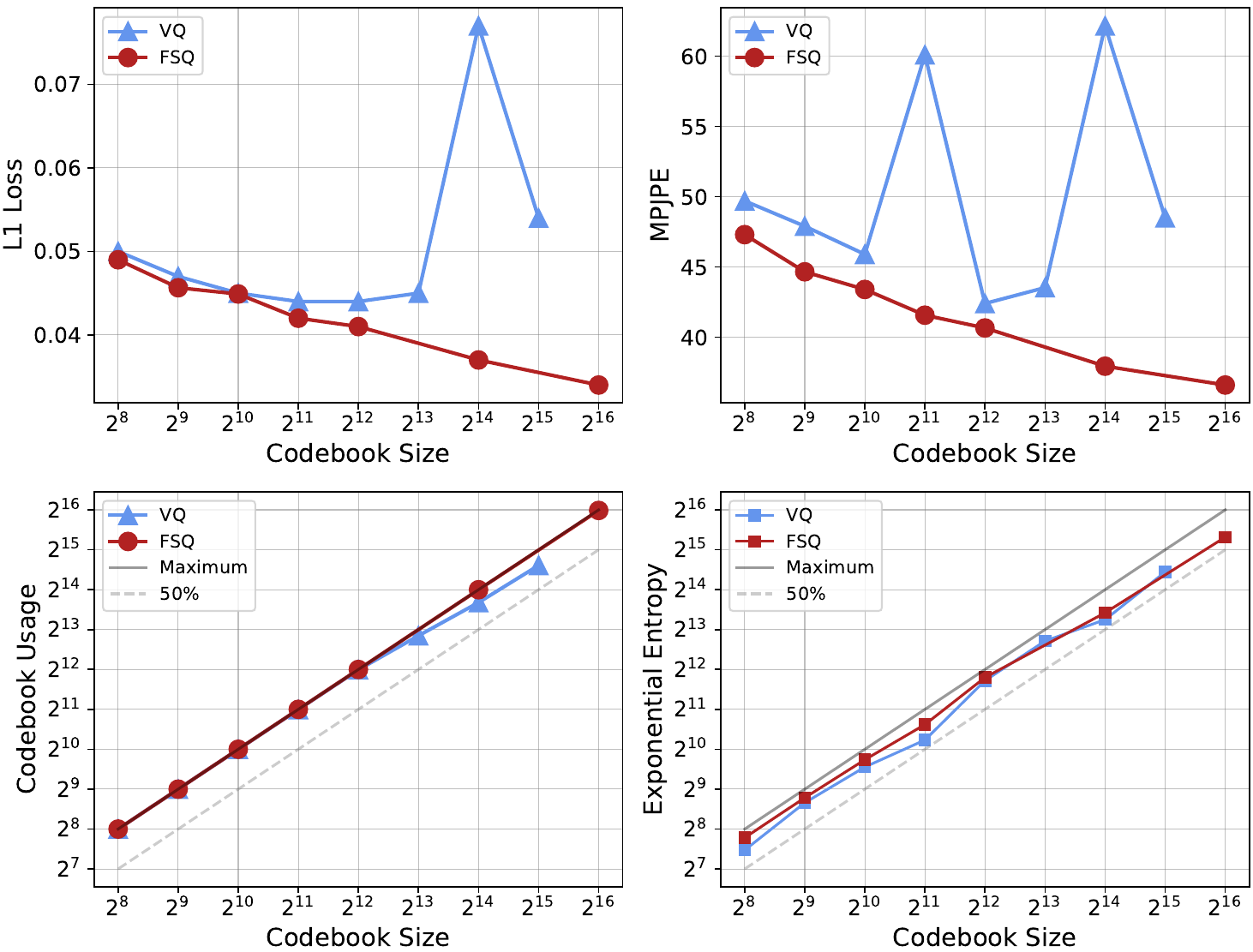}
		\caption{MotionUnion.}
	\end{subfigure}
    \vspace{-2em}
	\caption{Reconstuction results of different tokenizers on HumanML3D and MotionUnion. Reconstruction: L1 loss and MPJPE. Codebook Utilization: Codebook Usage and Entropy.}
    \vspace{-0.5em}
	\label{fig:tokenizer}
\end{figure}

\begin{figure*}
	\centering
    \vspace{-1em}
    \captionsetup[subfigure]{aboveskip=0pt, belowskip=-1em}
	\begin{subfigure}{0.329\linewidth}
		\centering
		\includegraphics[width=1.0\linewidth]{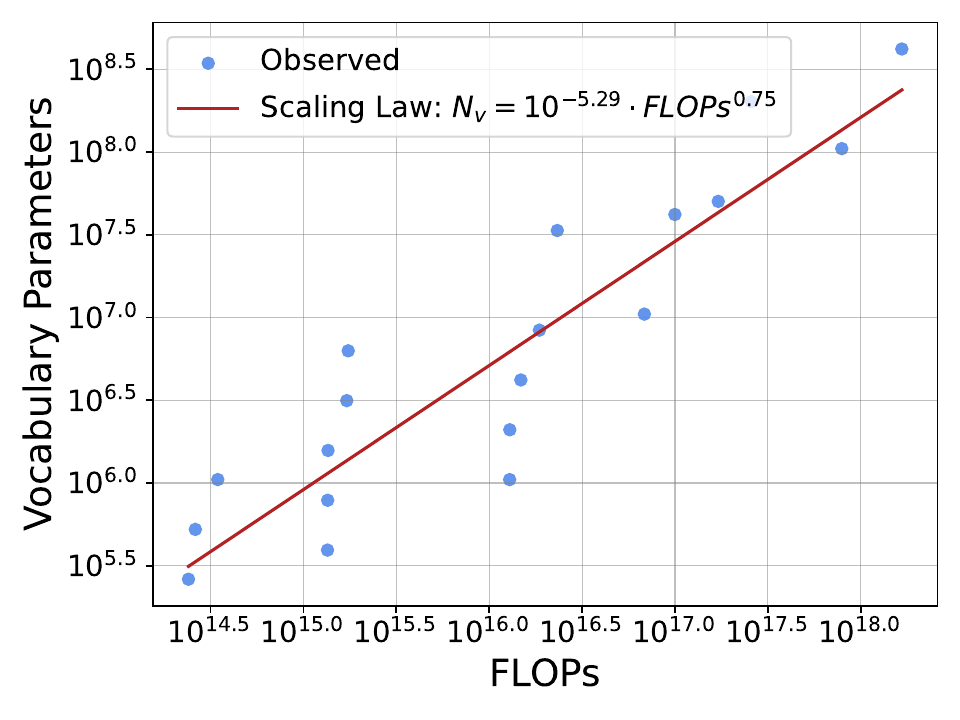}
		\caption{Vocabulary Parameter Scaling Law.}

	\end{subfigure}
	\begin{subfigure}{0.329\linewidth}
		\centering
		\includegraphics[width=1.0\linewidth]{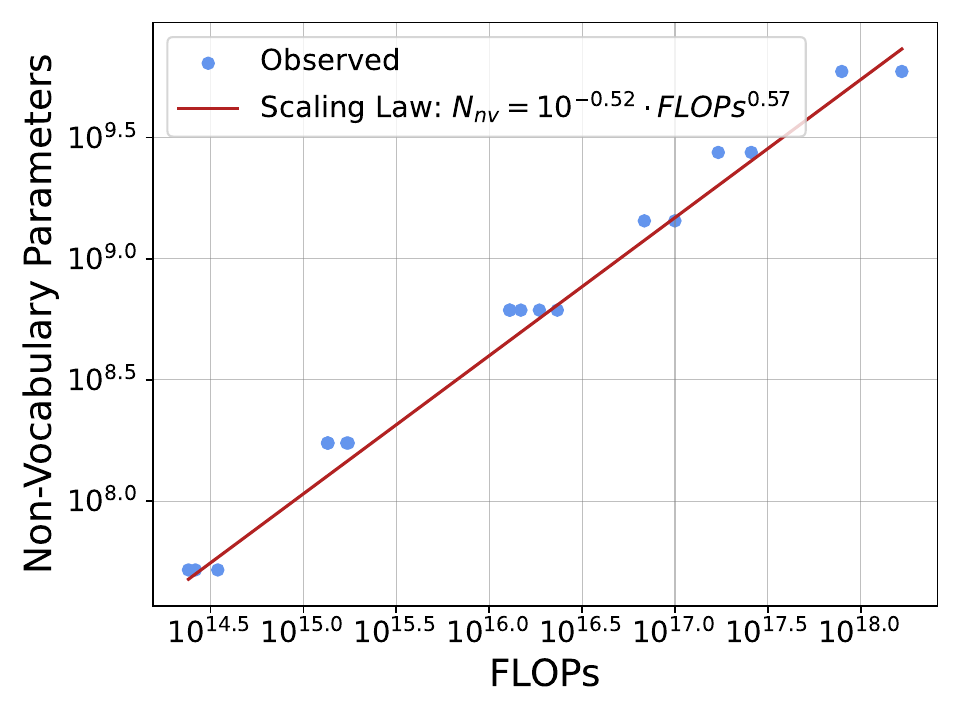}
		\caption{Non-Vocabulary Parameter Scaling Law.}
	\end{subfigure}
	\begin{subfigure}{0.329\linewidth}
		\centering
		\includegraphics[width=1.0\linewidth]{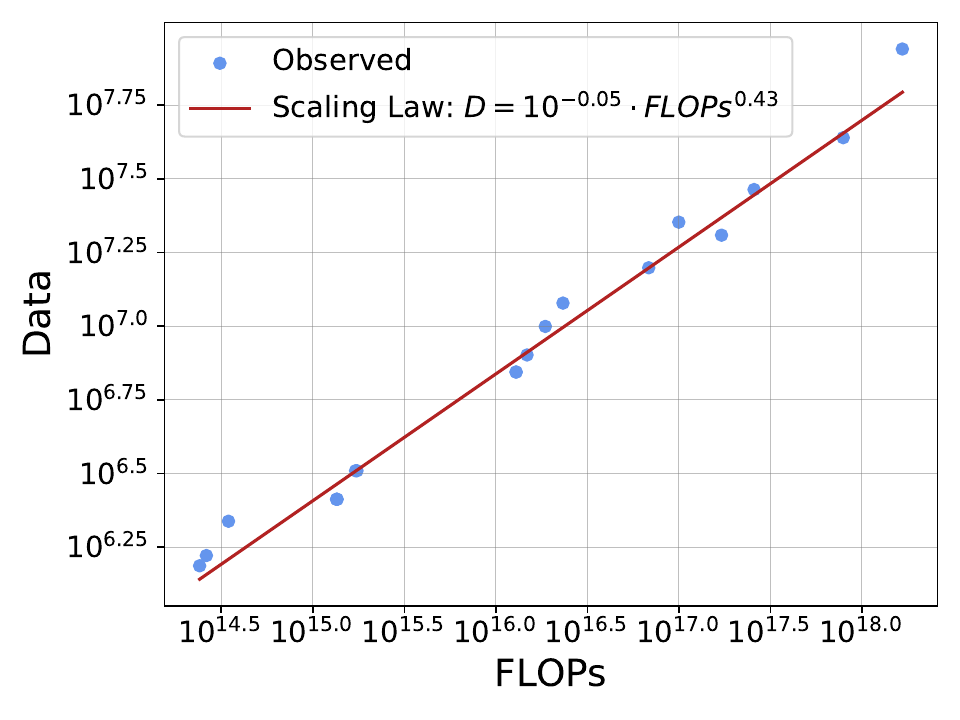}
		\caption{Data Scaling Law.}
	\end{subfigure}
    \vspace{-1em}
	\caption{Power laws of vocabulary parameters, non-vocabulary parameters, and data with respect to FLOPs.}
    \vspace{-1em}
	\label{fig:flops scaling law}
\end{figure*}

\begin{figure}[tb] \centering
   \includegraphics[width=1\linewidth]{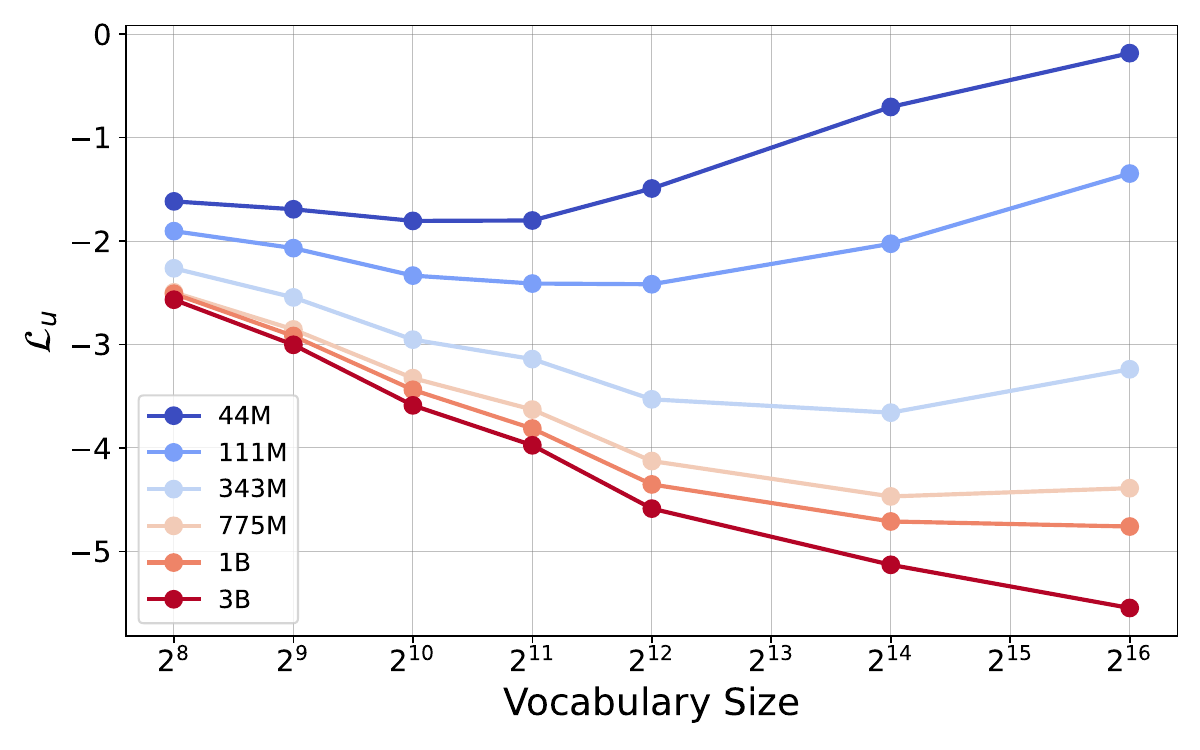}
    \vspace{-2.4em}
    \caption{$\mathcal{L}_u$ curves under various model and vocabulary sizes.} 
    \vspace{-1em}
\label{fig:vocab_size_loss_by_model}
\end{figure}

\section{Experiments}
\label{sec:experiments}
In our experiments, we aim to answer the following research questions and conduct corresponding ablation studies. 
\begin{itemize}[leftmargin=*]
    \item {
        \textbf{RQ1:} Is FSQ-VAE more effective and scalable than the vanilla VQ-VAE?
    }
    \item {
        \textbf{RQ2:} What model and vocabulary sizes should we use to get the best results with infinite computation resources?
    }
    \item {
        \textbf{RQ3:} Given a pre-defined compute budget in FLOPs, how should we choose the model size, the vocabulary parameters and how much data should we collect?
    }
    \item {
        \textbf{RQ4:} Can we predict the normalized loss based on a given computation budget, like FLOPs?
    }
\end{itemize}

\begin{table}[h]
\centering
\begin{adjustbox}{max width=1.\textwidth}
\begin{tabular}{lccc}
\toprule
\textbf{Parameters} & \textbf{Layers} & \textbf{Heads} & \textbf{Hidden size} \\
\midrule
ScaMo-44M  & 8  & 8  & 512  \\
ScaMo-111M & 12 & 12 & 768  \\
ScaMo-343M & 24 & 16 & 1024 \\
ScaMo-775M & 36 & 20 & 1280 \\
ScaMo-1.4B   & 48 & 24 & 1536 \\
ScaMo-3B   & 24 & 32 & 3200 \\
\bottomrule
\end{tabular}
\end{adjustbox}
\vspace{-0.7em}
\caption{Model Configuration Table}
\vspace{-2em}
\label{tab:model_config}
\end{table}

\subsection{Experiments Settings}
\myPara{Evaluation dataset}. 
We evaluate the motion tokenizer using HumanML3D and our MotionUnion. HumanML3D dataset contains 14,616 motions from AMASS~\citep{amass} and HumanAct12~\citep{action2motion}, and with along 44,970 descriptions. The evaluation part of MotionUnion is a held-out dataset from the original whole dataset and is divided into training, testing, and validation with a ratio of $80\%: 15\%: 5\%$.

\myPara{Implementation details.}

For the Motion FSQ-VAE, both the encoders and decoders are designed as convolutional residual blocks, utilizing a downsampling factor of 4. Regarding the prefix motion transformers, detailed specifications and configurations are provided in \cref{tab:model_config}, where layers indicate the number of transformer decoder blocks. The transformer architecture closely aligns with that of LLaMA. Specifically, each block incorporates RMSNorm prior to both the prefix attention layer and the feed-forward network (FFN) layer.
Due to space constraints, additional implementation details are presented in the appendix.

\myPara{Evaluation Metrics.}

We use reconstruction loss, mean per joint position error (\aka MPJPE), codebook utilization, and exponential entropy (\aka Entropy) to evaluate motion tokenizers. The reconstruction loss and MPJPE reflect the reconstruction performance, while the codebook utilization and Entropy reflect the percentage of codes used in the test set. What we are pursuing is the lower reconstruction loss and MPJPE. The ideal motion tokenizer should have almost full codebook utilization and a higher Entropy.
For the generation results, we conduct experiments on the HumanML3D benchmark. Following previous work, we use FID, R-precision, and matching score to evaluate the generation results. 
For validation of scaling law, we found FID is biased to the motion domain, and the pretrained motion feature extractor makes it hard to distinguish the differences between different models. Therefore, we follow previous work~\citep{scalingvocab} to use the normalized loss described in~\cref{eq:normalized_loss}.

\subsection{Experiments on Different Tokenizers (RQ1)}
\label{sec:5.2}

We present comprehensive experimental results across a range of codebook sizes, which are illustrated in~\cref{fig:tokenizer}. Numerical values are provided in the appendix. As shown in~\cref{fig:tokenizer}, the results clearly indicate that FSQ consistently outperforms VQ across various metrics on both the smaller HumanML3D dataset and the larger MotionUnion dataset. Notably, FSQ achieves comparable reconstruction performance to VQ with smaller codebook sizes. Additionally, as the codebook size increases, it performs much better on both reconstruction Loss and MPJPE than VQ, which suggests that FSQ can reconstruct data more accurately, particularly as the vocabulary size expands. This key observation demonstrates a clear advantage over VQ in terms of reconstruction fidelity. The benefits of FSQ become especially evident at larger codebook sizes, where it continues to maintain consistently lower error rates.

Moreover, FSQ exhibits the capacity to scale with larger codebook sizes while maintaining steadily increasing performance across datasets. As evidenced by the results in~\cref{fig:tokenizer}, the reconstruction Loss and MPJPE in FSQ steadily decrease with increasing codebook size, reflecting its robustness and stability when handling larger vocabularies. In contrast, the vanilla VQ exhibits instability, where its performance fluctuates significantly as the codebook size increases. This suggests that VQ struggles to achieve consistent performance when scaling to larger codebooks.

Additionally, FSQ demonstrates superior codebook utilization, even at very large codebook sizes. As can be seen in~\cref{fig:tokenizer}, the utilization of FSQ remains close to the theoretical maximum, indicating that it effectively leverages the available capacity of the codebook. This effective utilization stands in stark contrast to VQ, whose utilization rate decreases as the codebook size grows, thereby highlighting FSQ’s greater efficiency in handling larger codebooks.

Furthermore, FSQ achieves higher Entropy, indicating a more uniform distribution of its codes. This uniformity is crucial as it suggests that FSQ mitigates risks such as code collapse, which occurs when certain codes dominate the encoding process. The consistently higher Entropy values observed for FSQ demonstrate its ability to utilize codes evenly across the codebook, even as the codebook size expands. In comparison, VQ shows significant fluctuations in Entropy, indicating a less stable and uniform distribution.

In summary, the superior performance of FSQ in terms of reconstruction accuracy, codebook utilization, and code distribution uniformity positions it as a more robust and scalable alternative than VQ. This advantage is particularly beneficial in scenarios that require high-capacity encoding, such as large-scale motion data, where effective codebook utilization and precise reconstruction are paramount.

\subsection{Larger Model deserves Larger Vocabulary Sizes~(RQ2)}

When we have an infinite computation budget, how should we choose the vocabulary and model specifications? We plot the relationship between normalized test loss, vocabulary size, and model size in~\cref{fig:vocab_size_loss_by_model}. Accordingly, we have the following observations. 

\myPara{Larger models consistently outperform smaller ones across all codebook sizes.} This trend highlights the inherent advantage of larger models in capturing complex patterns, as they achieve lower normalized test losses compared to smaller ones. The improved performance of larger models across the vocabulary range implies that their enhanced capacity allows them to represent language features with greater fidelity, even when the vocabulary size changes. This observation aligns with the general expectation in deep learning that scaling up model parameters.

\myPara{Increasing vocabulary size requires a correspondingly larger model to fully utilize the potential of the large codebook size.} Smaller models struggle to maintain low test loss as the vocabulary grows, However, larger models, particularly those with billions of parameters, continue to perform well or even improve. This suggests a synergistic relationship where larger models are capable of leveraging the diversity and complexity afforded by a more extensive vocabulary. As a result, they can generalize better across diverse motion token inputs, while smaller models appear constrained by limited capacity and therefore fail to benefit as much from the increased vocabulary size.

\myPara{Large model deserves large vocabulary sizes.} The data also indicate that larger models actually benefit from larger vocabularies, as shown by their improved performance up to a certain vocabulary threshold, beyond which the benefits may taper off or slightly diminish. Smaller vocabularies might restrict large models' representational power, curtailing their ability to perform optimally on complex language tasks. In essence, large models "deserve" larger vocabularies, as this pairing allows them to reach their full potential.

\myPara{Power law between $N_v$ and $N_{nv}$.} \cref{fig:scaling law}~a demonstrates a strong power correlation between model parameters and vocabulary size, further underscores this relationship. The high \( R^2 \) value of 0.95 suggests that vocabulary size can be effectively predicted based on model size, 
\vspace{-0.5em}
\begin{equation}
    N_{v} = 10^{-5.604} \cdot N_{nv}^{1.467}.
    \vspace{-0.5em}
    \label{eq:nv_nnv}
\end{equation}
We predefine 3B models with different vocabulary sizes. We draw the best $N_v$ in ~\cref{fig:scaling law}~a with a yellow star. The yellow star can align with the fitted line, further validating this power law.
This law in~\cref{eq:nv_nnv} provides a practical tool to determine an optimal vocabulary size for a given transformer model size. We hope this power law can help the community choose the correct vocabulary size to get the best performance when using auto-regressive transformers. 

We answer RQ2 here. When we have no computation budgets, we should use a larger model and vocabulary size, but these should still adhere to the power law above.

\subsection{Power Law in Model Size, Vocabulary Size, and Data (RQ3)}
\label{sec:5.4}

As demonstrated in \cref{sec: scaling law in our model}, we extend the methodology proposed by Chinchilla~\citep{hoffmann2022training} by conducting a series of comprehensive experiments that explore variations in both model size and vocabulary size, as illustrated in \cref{fig:teaser_all_loss}. For each combination of model and vocabulary sizes, we identify the triplet \((N_{nv}, N_v, D)\) that minimizes the test loss. Subsequently, we visualize the relationship between model size, vocabulary size, and motion tokens with respect to FLOPs, as shown in \cref{fig:flops scaling law}. We hypothesize that these attributes follow a power-law relationship with FLOPs, expressed as \(N_{v}^{\text{opt}} \propto C^a, \quad N_{nv}^{\text{opt}} \propto C^b, \quad D^{\text{opt}} \propto C^c.\).

To fit this power-law model, we first determine \(C^c\) based on the data, subject to the constraint \((C^a + C^b) \cdot C^c = C.\) We then optimize the coefficients \(a\) and \(b\) corresponding to the optimal vocabulary size \(N_v^{\text{opt}}\) and the optimal non-vocabulary size \(N_{nv}^{\text{opt}}\). To empirically validate this hypothesis, we employ least squares estimation (LSE) to fit the model to the data. 

\myPara{Power law between $N_v$, $N_{nv}$, and $D$ with $C$.} 
From our estimation, the optimal model parameters and tokens scale with the compute budget is as, 
\vspace{-0.5em}
\begin{equation}
    N_{v} = 10^{-5.29} \cdot C^{0.75},
\end{equation}
\begin{equation}
    N_{nv} = 10^{-0.52} \cdot C^{0.57},
\end{equation}
\begin{equation}
    D = 10^{-0.05} \cdot C^{0.43},
\end{equation}
where $N_{nv}/D \propto C^{1.325} >C$ indicates we should scale $N_nv$ faster than data. Similarly, $N_{v}/N_{nv} \propto C^{1.315} >C$ indicates we should scale $N_v$ faster than $N_{nv}$.

\subsection{Scaling Law in FLOPs and The Test Loss (RQ4)}


\myPara{Logarithmic law in FLOPs and normalized test loss.}  Following the way of choosing representative points introduced in~\cref{sec:5.4}, we identify the optimal parameter combinations $(N_{nv}, N_{v}, D)$ that minimize test loss for a given FLOPs. The relationship between FLOPs and normalized test loss is depicted in~\cref{fig:scaling law}~(b).

We assume a logarithmic relationship between the computational cost $C$ and the normalized test loss $\mathcal{L}_u$, hypothesized as $\mathcal{L}_u \propto -\log_{10}(C)$. Same as ~\cref{sec:5.4}, we employ least squares estimation (\aka LSE) to fit a logarithmic curve to the data. The logarithmic scaling relationship can be expressed as,
\vspace{-0.7em}
\begin{equation}
    \mathcal{L}_u = -1.062 \times \log_{10}(C) + 13.839, 
    \vspace{-0.7em}
\end{equation}
which indicates that increasing the number of FLOPs consistently leads to a reduction in normalized test loss.

\myPara{Accurate prediction of normalized test loss using the logarithmic law.} To further validate the predictive accuracy of our proposed scaling law, we trained a large-scale model containing 3B parameters and a codebook size of $2^{16}$, resulting in a computational cost exceeding $10^{18}$ FLOPs. As shown in~\cref{fig:flops scaling law}~(a), the fitted logarithmic curve closely aligns with the observed test loss values, thereby substantiating the robustness and generalizability of our scaling law for predicting model performance.

\subsection {Ablation Studies}
\begin{table}[]
\vspace{0.5em}
\begin{adjustbox}{max width=0.48\textwidth}
\begin{tabular}{cc|ccc}
\toprule
Text Enc. & Prefix & FID $\downarrow$    & Matching Score $\uparrow$ & Top1 R-P $\uparrow$  \\ 
\midrule
GT           &    -    &   -     & 2.974          & 0.511 \\ 
CLIP         &   $\times$     & 0.226 & 3.422           & 0.402  \\ 
T5-XL        &   $\checkmark$     & \textbf{0.104} & \textbf{3.021}           & \textbf{0.510} \\ 
\bottomrule
\end{tabular}
\end{adjustbox}
\vspace{-0.8em}
\caption{Ablation experiments of the architecture.}
\vspace{-1em}
\label{tab: ablation}
\end{table}

We conduct an ablation study on a scalable architecture by training two distinct model variants, both with a parameter size of 343M, consistent with the T2M-GPT configuration \citep{t2mgpt}. The first variant employs a CLIP text encoder without the text-prefix design, while the other utilizes a T5-XL text encoder with text-prefixed design. Both models are trained on our constructed dataset MotionUnion and evaluated on the HumanML3D benchmark. As shown in \cref{tab: ablation}, the model with T5-XL and prefix design achieves a substantial performance improvement, with the FID decreasing from 0.226 to 0.104, the matching score increasing from 3.422 to 3.021, and the Top-1 R-Precision improving from 0.402 to 0.510. These results underscore the effectiveness of the text-prefixed autoregressive transformer architecture.

\section{Conclusion}
\label{sec:conclusion}

In this paper, we present a scalable text-driven motion generation system, comprising a motion FSQ-VAE and a text-prefix autoregressive transformer. Additionally, we provide the first empirical analysis of scaling behaviors within the motion generation domain. Our experiments reveal a logarithmic relationship between the compute budgets in FLOPs and normalized test loss. Furthermore, we observe that both vocabulary size and non-vocabulary size, as well as the amount of data, exhibit power-law dependencies with respect to FLOPs. We also identify a power-law relationship between vocabulary size and non-vocabulary size. We hope that our findings will offer valuable insights for future research and practical applications in the field of motion synthesis.
{
    \small
    \bibliographystyle{ieeenat_fullname}
    \bibliography{main}
}

\appendix
\clearpage
\etocdepthtag.toc{mtappendix}
\etocsettagdepth{mtchapter}{none}
\etocsettagdepth{mtappendix}{subsection}
\setcounter{page}{1}
\maketitlesupplementary
\onecolumn

\renewcommand{\lstlistingname}{Code}

\section{Flops calculation}
\label{sec:rationale}
%

In this paper, we follow the criteria proposed by OpenAI~\citep{gpt4} to calculate the floating point operations (FLOPs). The detailed breakdown of these computations is provided in~\cref{tab:FLOPs}. Additionally, to facilitate reproducibility, we present the complete code for this calculation process in Code~\ref{flops}.


By utilizing the provided code, we can quickly compute the FLOPs for their specific model configurations, facilitating performance analysis and design optimization.

\begin{table}[h]
\centering
\begin{adjustbox}{max width=1\textwidth}
\begin{tabular}{l|c|c}
\toprule[1.2pt]
\textbf{Operation} & \textbf{Parameters} & \textbf{FLOPs per Token} \\
\midrule
Embed & $(n_{\text{vocab}} + n_{\text{ctx}}) d_{\text{model}}$ & $4 d_{\text{model}}$ \\

Attention: Q, K, V & $n_{\text{layer}} \times d_{\text{model}}\times  3 d_{\text{attn}}$ & $2 n_{\text{layer}} \times d_{\text{model}} \times 3 d_{\text{attn}}$ \\

Attention: Mask & -- & $2 n_{\text{layer}} \times n_{\text{ctx}}\times  d_{\text{attn}}$ \\

Attention: Projection & $n_{\text{layer}} \times d_{\text{attn}}\times d_{\text{model}}$ & $2 n_{\text{layer}} \times d_{\text{attn}} \times d_{\text{model}}$ \\

Feedforward & $n_{\text{layer}}\times  2 d_{\text{model}} \times d_{\text{ff}}$ & $2 n_{\text{layer}} \times 2 d_{\text{ff}}$ \\

De-embed & -- & $2 d_{\text{model}} \times n_{\text{vocab}}$ \\

\textbf{Total (Non-Embedding)} & $N = 2 d_{\text{model}}\times  n_{\text{layer}} (2 d_{\text{attn}} + d_{\text{ff}})$ & $C_{\text{forward}} = 2 N + 2 n_{\text{layer}} \times n_{\text{ctx}}\times  d_{\text{attn}}$ \\
\bottomrule[1.2pt]
\end{tabular}
\end{adjustbox}
\caption{Details of FLOPs calculation criteria by OpenAI~\cite{gpt4}.}
\label{tab:FLOPs}
\end{table}

\lstset{style=pythonstyle}
\begin{lstlisting}[language=Python, caption=Open AI method for forward pass FLOPs counting of decoder-only Transformer., label=flops]
def openai_flops_per_token(n_layers, n_heads, d_model, n_ctx, n_vocab, ff_ratio=4):
    """Open AI method for forward pass FLOPs counting of decoder-only Transformer
    """
    d_attn = d_model // n_heads
    d_ff = d_model * ff_ratio
 
    embeddings = 4 * d_model
    attn_qkv = 2 * n_layers * d_model * 3 * (d_attn * n_heads)
    attn_mask = 2 * n_layers * n_ctx * (d_attn * n_heads)
    attn_project = 2 * n_layers * (d_attn * n_heads) * d_model
    ff = 2 * n_layers * 2 * d_model * d_ff
    logits = 2 * d_model * n_vocab
 
    return embeddings + attn_qkv + attn_mask + attn_project + ff + logits
\end{lstlisting}

\section{Implmentation Details}

The framework is implemented using PyTorch. For the Motion FSQ-VAE, both the encoders and decoders are designed as convolutional residual blocks, utilizing a downsampling factor of 4. The transformer architecture closely aligns with that of LLaMA. Specifically, each block incorporates RMSNorm prior to both the prefix attention layer and the feed-forward network (FFN) layer. We train the transformers using bf16 to reduce the memory. We do not use the masking strategy in~\citep{t2mgpt}. The optimization details are shown in~\cref{tab:optimizer}. 

\begin{table}[h]
\centering
\begin{adjustbox}{max width=1\textwidth}
\begin{tabular}{l|c|c}
\toprule[1.2pt]
Config                 & Tokenizer   & Transformer             \\ \midrule
optimizer              & AdamW       & AdamW                   \\
optimizer momentum     & 0.9         & 0.9                     \\
weight decay           & 0.0         & 1e-06                   \\
learning rate schedule & MultiStepLR & Warmup and Cosine decay \\
milestone\_ratio       & 0.6         & -                       \\
warmup ratio           & 0.003       & 0.1                     \\ \bottomrule[1.2pt]
\end{tabular}
\end{adjustbox}
\caption{The optimizer details.}
\label{tab:optimizer}
\end{table}

For motion representation, we follow HumanML3D~\citep{t2mgpt}. \textbf{HumanML3D Format} proposes a motion representation $x^{1:L}$ inspired by motion features in character control. This redundant representation is quite suited to neural models, particularly variational autoencoders. Specifically, the $i$-th pose $x^i$ is defined by a tuple of root angular velocity $r^a \in \mathbb{R}$ along Y-axis, root linear velocities ($r^x, r^z \in \mathbb{R}$) on XZ-plane, root height $r^y \in \mathbb{R}$, local joints positions $j^p \in \mathbb{R}^{3N_j}$, velocities $j^v \in \mathbb{R}^{3N_j}$ and rotations $j^r \in \mathbb{R}^{6N_j}$ in root space, and binary foot-ground contact features $c^f \in \mathbb{R}^4$ by thresholding the heel and toe joint velocities, where $N_j$ denotes the joint number, giving:
\[
x^i = \{r^a, r^x, r^z, r^y, j^p, j^v, j^r, c^f\}.
\]


\section{Tokenizer Results}
We show the numerical results of different tokenizers here. The superior performance of FSQ in terms of reconstruction accuracy, codebook utilization, and code distribution uniformity positions it as a more robust and scalable alternative than VQ. This advantage is particularly beneficial in scenarios that require high-capacity encoding, such as large-scale motion data, where effective codebook utilization and precise reconstruction are paramount.

\begin{table}[htbp]
    \centering
    \begin{subtable}{0.45\linewidth}
        \centering
                    \resizebox{0.99\textwidth}{!}{
        \begin{tabular}{l|c|c|c|c|c}
            \toprule[1.2pt]
            VQ    & L1 loss & FID     & MPJPE   & Activate &  Entropy                           \\ \midrule
            256   & 0.071  & 0.12 & 0.05 & 1.00        & 170.58 \\
            512   & 0.065 & 0.10     & 0.05  & 1.0        & 364.45                        \\
            1024  & 0.062 & 0.09 & 0.05 & 0.99   & 704.70                         \\
            2048  & \textbf{0.060} & \textbf{0.08}  & \textbf{0.05} & 0.97    & 1145.77                         \\
            4096  & 0.078 & 0.78  & 0.09   & 0.69   & 147.96                         \\
            8192  & 0.083  & 21.8  & 0.17   & 0.81     & 82.20                            \\
            32768 & 0.076 & 5.05    & 0.11   & 0.63    & 1029.20                          \\ \bottomrule[1.2pt]
        \end{tabular}
        }
        \caption{VQ results on HumanML3D.}
    \end{subtable}
    \hfill 
    \begin{subtable}{0.45\linewidth}
        \centering
        \setlength{\tabcolsep}{11pt}
                    \resizebox{0.99\textwidth}{!}{
        \begin{tabular}{l|c|c|c|c}
            \toprule[1.2pt]
            VQ    & L1 loss & MPJPE & Activate & Entropy       \\ \midrule
            256   & 0.050    & 49.70  & 1.00        & 177.65    \\
            512   & 0.047   & 47.90  & 1.00        & 404.09   \\
            1024  & 0.045   & 45.90  & 0.998    & 752.028   \\
            2048  & 0.044   & 60.10  & 0.996    & 1202.23   \\
            4096  & \textbf{0.044}   & \textbf{42.40}  & 0.994   & 3373.35   \\
            8192  & 0.045   & 43.53 & 0.998   & 6714.25  \\
            16384 & 0.077   & 62.15 & 0.993    & 12732.29  \\
            32768 & 0.054   & 48.50  & 0.962   & 22286.32 \\ \bottomrule[1.2pt]
        \end{tabular}
        }
        \caption{VQ results on MotionUnion.}
    \end{subtable}
    \\[1cm] 
    \begin{subtable}{0.45\linewidth}
        \centering
                    \resizebox{0.99\textwidth}{!}{
        \begin{tabular}{l|c|c|c|c|c}
        \toprule[1.2pt]
            FSQ   & L1 loss & FID    & MPJPE   & Activate & Entropy        \\ \midrule
            256   & 0.081   & 0.159  & 0.057  & 1.00        & 213.20      \\
            512   &  0.075  & 0.129  & 0.053  & 1.00        & 446.10      \\
            1024  & 0.0713  & 0.106  & 0.052  & 1.00        & 723.80      \\
            4096  & 0.064   & 0.088  & 0.049 & 0.998    & 2759.52    \\
            16384 & 0.053   & 0.052 & 0.044  & 0.976    & 10119.25  \\
            65536 & \textbf{0.051}  & \textbf{0.049}  & \textbf{0.042}   & 0.764    & 33818.21 \\ \bottomrule[1.2pt]
        \end{tabular}
        }
        \caption{FSQ results on HumanML3D.}
    \end{subtable}
    \hfill 
    \begin{subtable}{0.45\linewidth}
        \setlength{\tabcolsep}{11pt}
        \centering
                    \resizebox{0.99\textwidth}{!}{
        \begin{tabular}{l|c|c|c|c}
            \toprule[1.2pt]
            FSQ   & L1 loss & MPJPE  & Activate & Entropy        \\ \midrule
            256   & 0.049   & 47.30   & 1.00        & 220.26   \\
            512   & 0.046 & 44.66 & 1.00        & 441.05    \\
            1024  & 0.045  & 43.40   & 1.00        & 853.64     \\
            2048  & 0.042   & 41.57  & 1.00        & 1572.82    \\
            4096  & 0.041   & 40.66  & 1.00        & 3561.95    \\
            16384 & 0.037   & 37.94  & 0.999   & 10974.16   \\
            65536 & \textbf{0.034}   & \textbf{36.60}   & 0.999     & 40818.21 \\ \bottomrule[1.2pt]
        \end{tabular}
        }
        \caption{FSQ results on MotionUnion.}
    \end{subtable}
    \caption{Tokenizer numerical results. The Entropy is Exponential Entropy.}
\end{table}


\section{More Results on HumnaML3D Benchmark}
We train evaluate their performance of the models on the HumanML3D benchmark. The numerical results are presented in \cref{tab:all human results}. Notably, we observe that the model configured with the largest codebook size and model capacity achieves the best overall performance, consistent with the lowest normalized test loss. However, when examining cases of overfitting—such as the combination of a small codebook size (e.g., 256) and a large model size (44M parameters)—the automatic metrics continue to improve, despite being inconsistent with the normalized loss. A similar phenomenon is observed when training T2M-GPT~\citep{t2mgpt}. We hypothesize that this discrepancy arises from the suboptimal performance of the pretrained feature extractor. Additionally, our findings suggest that larger codebook sizes necessitate proportionally larger model capacities to fully leverage their potential.

Furthermore, we conduct a comparative analysis against other frameworks that directly fine-tune large language models (LLMs), such as those proposed in~\citep{wang2024quo, motiongpt, wu2024motionllm, avatargpt}. Our approach demonstrates competitive results on semantic alignment metrics, including R@1, R@3, and Matching Score. Notably, our model achieves superior performance in terms of the FID, highlighting the advantages of our motion tokenizer and architectural design. These results indicate that training a native motion generation model from scratch offers substantial benefits compared to fine-tuning an LLM. Specifically, this approach not only improves performance but also achieves significant parameter efficiency.

\begin{table}[htbp]
\centering
\begin{adjustbox}{max width=1\textwidth}
\begin{tabular}{ll|cccccc}

\toprule[1.2pt]
Model & Model Size & FID$\downarrow$   & R@1$\uparrow$    & R@2$\uparrow$    & R@3$\uparrow$    & Matching Score$\downarrow$  & Diversity \\ \midrule
MotionGPT~\citep{motiongpt}* & Llama-1-13B & 0.592 & 0.363 & - & 0.633 & 4.029 & - \\
MotionGPT~\citep{motiongpt}* & Llama-2-13B & 0.571 & 0.367 & - & 0.654 & 3.981 & - \\
MotionLLM~\citep{wu2024motionllm} & Gemma-2b & 0.491 & 0.482 & - & 0.770 & 3.138 & - \\
AvatarGPT~\citep{avatarclip} & Llama-1-13B & 0.567 & 0.389 & - & 0.623 & - & - \\
LargeMotionModel~\citep{wang2024quo} & Llama-2-13B & 0.166 & 0.519 & - & 0.803 & 2.964 & -\\
\midrule
Codebook size & Model Size & FID$\downarrow$   & R@1$\uparrow$    & R@2$\uparrow$    & R@3$\uparrow$    & Matching Score$\downarrow$  & Diversity \\ \midrule

256           & 44M        & 3.184   & 0.302  & 0.45   & 0.547  & 4.557          & 8.317    \\
256           & 111M       & 1.197   & 0.398  & 0.565  & 0.667  & 3.726          & 8.968     \\
256           & 343M       & 0.730    & 0.432  & 0.618  & 0.719  & 3.466          & 8.972     \\
256           & 775M       & 0.704   & 0.434  & 0.617  & 0.722  & 3.428          & 9.393     \\
256           & 1B         & 0.709   & 0.441  & 0.626  & 0.723  & 3.424          & 9.123     \\
256           & 3B         & 0.670    & 0.443  & 0.627  & 0.726  & 3.410           & 8.738     \\
\midrule
512           & 44M        & 3.971   & 0.271 & 0.402  & 0.498  & 4.981          & 8.792     \\
512           & 111M       & 1.338   & 0.373  & 0.550   & 0.660   & 3.741          & 8.567     \\
512           & 343M       & 0.851   & 0.415  & 0.590   & 0.695  & 3.514          & 9.226     \\
512           & 775M       & 0.664   & 0.441  & 0.619  & 0.727  & 3.361          & 9.187    \\
512           & 1B         & 0.624   & 0.447 & 0.631 & 0.734  & 3.330           & 8.948     \\
512           & 3B         & 0.617   & 0.443 & 0.627  & 0.734  & 3.340         & 9.217     \\
\midrule
1024          & 44M        & 8.111   & 0.216  & 0.332  & 0.415  & 5.766          & 7.614    \\
1024          & 111M       & 1.331   & 0.371 & 0.535 & 0.647 & 3.865         & 9.118    \\
1024          & 343M       & 0.815   & 0.422  & 0.601  & 0.705  & 3.525          & 9.404     \\
1024          & 775M       & 0.583   & 0.447 & 0.635  & 0.735 & 3.300            & 9.489     \\
1024          & 1B         & 0.488   & 0.453  & 0.650   & 0.745  & 3.290           & 9.136     \\
1024          & 3B         & 0.496   & 0.453 & 0.643 & 0.741 & 3.296          & 9.376     \\
\midrule
2048          & 44M        & 13.964  & 0.192 & 0.298 & 0.372  & 6.295          & 6.548     \\
2048          & 111M       & 1.553   & 0.361  & 0.528  & 0.640   & 3.857          & 9.11      \\
2048          & 343M       & 0.794   & 0.418 & 0.604  & 0.707  & 3.490         & 9.136     \\
2048          & 775M       & 0.465   & 0.450   & 0.636  & 0.736  & 3.300            & 9.241     \\
2048          & 1B         & 0.320    & 0.454  & 0.640   & 0.740   & 3.264         & 9.836     \\
2048          & 3B         & 0.346   & 0.465 & 0.656 & 0.752 & 3.216         & 9.277     \\
\midrule
4096          & 44M        & 18.311  & 0.131  & 0.217  & 0.276  & 7.077         & 6.043     \\
4096          & 111M       & 1.465  & 0.327 & 0.492 & 0.599 & 4.134          & 8.542    \\
4096          & 343M       & 0.568  & 0.422 & 0.587  & 0.689  & 3.467          & 9.174     \\
4096          & 775M       & 0.240    & 0.464  & 0.650   & 0.750   & 3.250           & 9.393     \\
4096          & 1B         & 0.208   & 0.486 & 0.672 & 0.771 & 3.120           & 9.564     \\
4096          & 3B         & 0.214   & 0.483  & 0.674  & 0.764  & 3.128          & 9.455     \\
\midrule
16384         & 44M        & 44.240   & 0.056  & 0.103  & 0.153  & 8.019          & 2.842     \\
16384         & 111M       & 4.714   & 0.254 & 0.395 & 0.496 & 4.891          & 8.030      \\
16384         & 343M       & 1.217  & 0.380 & 0.556  & 0.661 & 3.711          & 8.838     \\
16384         & 775M       & 0.501   & 0.443  & 0.625  & 0.723  & 3.370           & 9.342     \\
16384         & 1B         & 0.347   & 0.477  & 0.657  & 0.758  & 3.206          & 9.727     \\
16384         & 3B         & 0.331   & 0.469  & 0.670   & 0.761  & 3.192          & 9.310      \\
\midrule
65536         & 44M        & 50.796 & 0.041 & 0.0791 & 0.1185 & 8.203          & 1.490      \\
65536         & 111M       & 2.178  & 0.311 & 0.461 & 0.566 & 4.286          & 5.311     \\
65536         & 343M       & 0.104   & 0.510   & 0.692  & 0.781 & 3.021          & 9.540      \\
65536         & 775M       & 0.150    & 0.495  & 0.685  & 0.785  & 3.080           & 9.558    \\
65536         & 1B         & 0.131    & 0.503  & 0.687  & 0.779  & 3.070           & 9.580      \\
65536         & 3B         & \textbf{0.101}   & \textbf{0.512}  & \textbf{0.695} & \textbf{0.796} & \textbf{2.990}           & 9.590      \\ \bottomrule[1.2pt]
\end{tabular}
\end{adjustbox}
\caption{Test results of different models on HumanML3D Benchmark. We take the results of MotionGPT* from Wang \etal~\citep{wang2024quo}.}
\label{tab:all human results}
\end{table}

\clearpage

\section{Dataset Visualization}
We show motion visualizations and text annotations of MotionUnion in~\cref{fig:dataset}. Render videos can be found in the supplementary materials. The specific frames and sequences are shown in the~\cref{tab:dataset frames}. PhysHumanML3D subset is the physics-optimized version of HumanML3D using HPC~\citep{phc}.

\begin{table}[htbp]
\centering
\vspace{-0.8em}
\begin{adjustbox}{max width=1\textwidth}
\begin{tabular}{l|ll}
\toprule[1.2pt]
                  & Frames  & Seqs  \\ \midrule
PhysHumanML3D & 5770156 & 22628 \\
Animation         & 55282   & 559   \\
Combatmotion      & 3368986 & 26097 \\
EgoBody           & 437976  & 980   \\
Fitness           & 106537  & 262   \\
Game Motion       & 797824  & 3296  \\
Haa500            & 438733  & 6944  \\
HumanML3D           & 4117392 & 29228 \\
Humman            & 187580  & 971   \\
Idea400           & 2108727 & 12042 \\
Kungfu            & 311507  & 1032  \\
Music             & 914642  & 3394  \\
Perform           & 327903  & 923   \\
100 Style         & 4018110 & 16074 \\
Internal Data     & 3905243 & 23067 \\ \bottomrule[1.2pt]
\end{tabular}
\end{adjustbox}
\vspace{-0.8em}
\caption{The detailed quantities of frames and sequences within the MotionUnion dataset.}
\vspace{-1.8em}
\label{tab:dataset frames}
\end{table}

\begin{figure*}[h]
    \centering
    \includegraphics[width=1\linewidth]{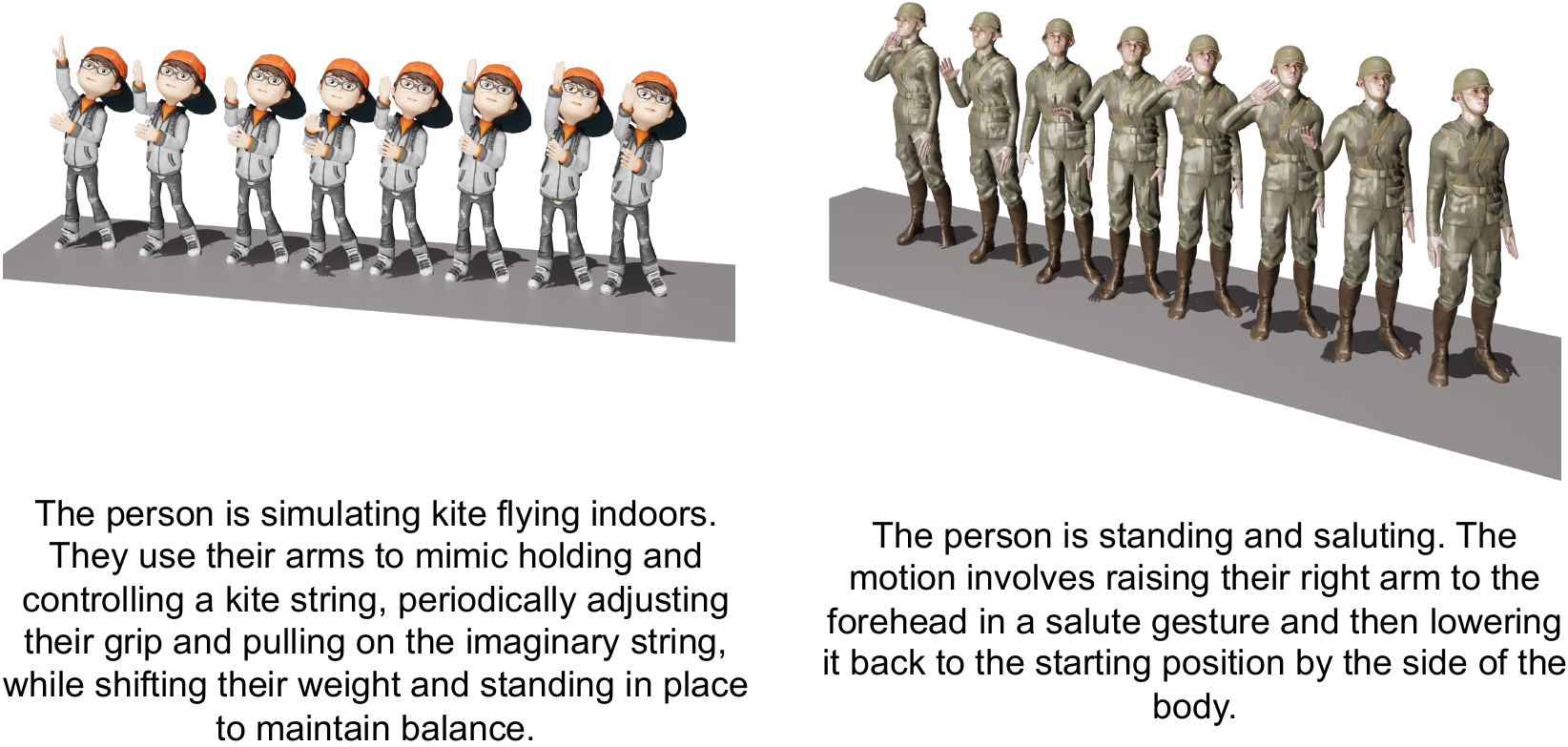}
    \vspace{-1.8em}
    \caption{MotionUnion visualization}
    \vspace{-1em}
    \label{fig:dataset}
\end{figure*}


\section{FSQ settings}
We follow Mentzer \etal~\citep{FSQ} to set the $L$ in~\cref{tab:FSQ L}. The codebook size can be calculated as $|\mathcal{C}| = \prod_{i=1}^{d} L_i$. For example, $2^{10} \approx 8 * 5*5*5 = 1000$.

\begin{table}[htbp]
    \centering
    \begin{adjustbox}{max width=1\textwidth}
    \begin{tabular}{cccccccccc}
        \toprule[1.2pt]
        Target size $|\mathcal{C}|$ & $2^4$ & $2^6$ & $2^8$ & $2^9$ & $2^{10}$ & $2^{11}$ & $2^{12}$ & $2^{14}$ & $2^{16}$ \\ 
        \midrule
        Quantized integer layers $L$ & $[5, 3]$ & $[8, 8]$ & $[8, 6, 5]$ & $[8, 8, 8]$ & $[8, 5, 5, 5]$ & $[8, 8, 6, 5]$ & $[7, 5, 5, 5,5]$ & $[8, 8, 8, 6, 5]$ & $[8, 8, 8, 5, 5, 5]$ \\ 
        \bottomrule[1.2pt]
    \end{tabular}
    \end{adjustbox}
    \vspace{-1em}
    \caption{The choices of $L$ in FSQ.}
    \vspace{-2em}
    \label{tab:FSQ L}
\end{table}

\clearpage

\section{More Generation Visualizations}
\begin{figure*}[h]
    \centering
    \includegraphics[width=0.93\linewidth]{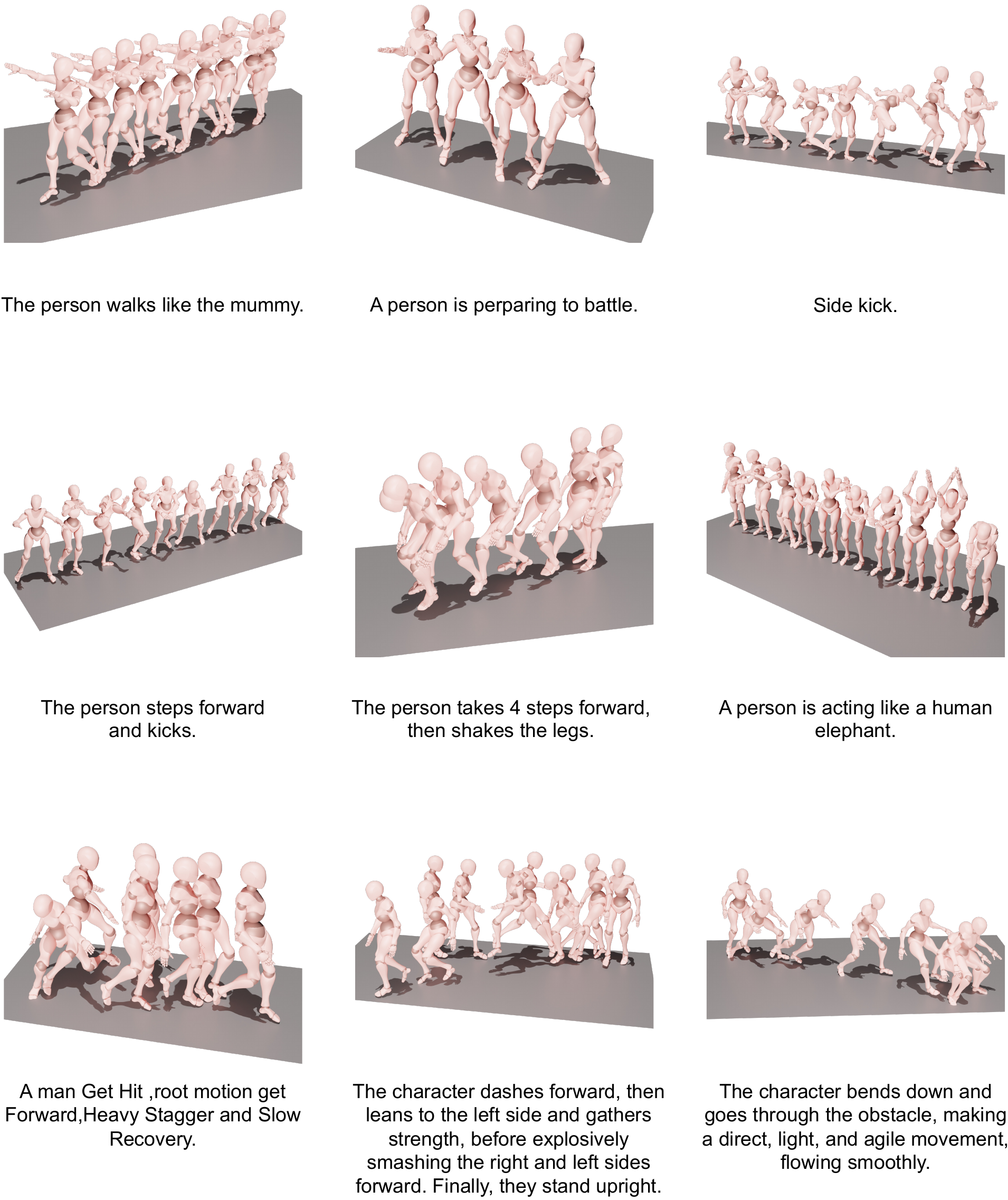}
    \caption{Human motion generation results on test set of ScaMo.}
    \label{fig:main-vis-gen}
\end{figure*}

We show some of the generation results in~\cref{fig:main-vis-gen}. The visualization shows our model could handle various types of texts. More generation visualizations and comparisons between different model sizes and codebook sizes can be found in the supplementary materials.

\clearpage

\section{Limitations}
The main limitation of this paper is the limited data. Unfortunately, we still have not observed the emerging abilities, based on these limited data. We are still working on collecting larger text-motion datasets and leaving it as our future work. Additionally, some of the data were sourced from video motion capture, which has posed quality constraints that, in turn, impact the generation quality.

\end{document}